\documentclass[
twocolumn,
]{ceurart}

\usepackage{listings}
\usepackage{tabularx}
\usepackage{booktabs}
\usepackage{tipa}
\usepackage{subcaption}
\usepackage{soul}
\usepackage{xcolor}

\newcommand{\ap}[1]{{\textcolor{black}{#1}}}
\newcommand{\mr}[1]{{\textcolor{black}{#1}}}
\newcommand{\apx}[1]{{\textcolor{black}{#1}}}

\newcommand{\last}[1]{{\textcolor{black}{#1}}}
\newcolumntype{L}[1]{>{\raggedright\let\newline\\\arraybackslash\hspace{0pt}}m{#1}}

\setuldepth{test}
\begin{document}

\copyrightyear{2024}
\copyrightclause{Copyright for this paper by its authors.
  Use permitted under Creative Commons License Attribution 4.0
  International (CC BY 4.0).}

\conference{CLiC-it 2024: Tenth Italian Conference on Computational Linguistics, Dec 04 — 06, 2024, Pisa, Italy}

\title{GFG - Gender-Fair Generation: \\ A CALAMITA Challenge}

\author[1,2]{Simona Frenda}[%
orcid=0000-0002-6215-3374,
email=s.frenda@hw.ac.uk,
]
\cormark[1]
\fnmark[1]
\address[1]{Interaction Lab, Heriot-Watt University, Edinburgh, Scotland}
\address[2]{aequa-tech, Turin, Italy}
\address[3]{Fondazione Bruno Kessler, Trento, Italy}
\address[4]{University of Trento, Trento, Italy}
\address[5]{Computer Science Department, University of Turin, Turin, Italy}
\address[6]{Universitat de Barcelona, Barcelona, Spain}
\address[7]{MaiNLP \& MCML, LMU Munich, Germany}

\author[3,4]{Andrea Piergentili}[%
orcid=0000-0003-2117-1338,
email=apiergentili@fbk.eu,
]
\cormark[1]
\fnmark[1]

\author[3]{Beatrice Savoldi}[%
orcid=0000-0002-3061-8317,
email=bsavoldi@fbk.eu,
]

\author[5]{Marco Madeddu}[%
orcid=0009-0004-5620-0631,
email=marco.madeddu@unito.it,
]

\author[6]{Martina Rosola}[%
orcid= 0000-0002-8891-352X,
email= martina.rosola@gmail.com,
]

\author[7]{Silvia Casola}[%
orcid= 0000-0002-0017-2975,
email= s.casola@lmu.de,
]

\author[5]{Chiara Ferrando}[%
email= chiara.ferrando@unito.it,
]

\author[5]{Viviana Patti}[%
orcid= 0000-0001-5991-370X,
email= viviana.patti@unito.it,
]

\author[3]{Matteo Negri}[%
orcid=0000-0002-8811-4330,
email=negri@fbk.eu,
]

\author[3]{Luisa Bentivogli}[%
orcid=0000-0001-7480-2231,
email=bentivo@fbk.eu,
]

\cortext[1]{Corresponding authors.}
\fntext[1]{These authors contributed equally.}

\begin{abstract}
  Gender-fair language aims at promoting gender equality by using terms and expressions that include all identities and avoid reinforcing gender stereotypes. Implementing gender-fair strategies is particularly challenging in heavily gender-marked languages, such as Italian. To address this, the Gender-Fair Generation challenge intends to help 
  shift toward
  gender-fair language in written communication. The challenge, designed to assess and monitor the recognition and generation of gender-fair language in both mono- and cross-lingual scenarios, includes three tasks: (1) the detection of 
  gendered expressions in Italian sentences, (2) the reformulation of 
  gendered expressions into gender-fair alternatives, and (3) the generation of gender-fair language in automatic translation from English to Italian. 
  The challenge relies on three different annotated datasets: the GFL-it corpus, which contains Italian texts extracted from administrative documents provided by the University of Brescia; GeNTE, a bilingual test set for gender-neutral rewriting and translation built
  upon
  a subset of the Europarl dataset; and Neo-GATE, a bilingual test set designed to assess the use of non-binary neomorphemes in Italian for both fair formulation and translation tasks.
 Finally, each task is evaluated with specific metrics: average of F1-score obtained by means of BERTScore computed on each entry of the datasets for task 1, an accuracy measured with a gender-neutral classifier, and 
 a coverage-weighted accuracy for tasks 2 and 3. 
  
\end{abstract}


\maketitle

\section{Challenge: Introduction and Motivation}

Gender-fair language, also known as inclusive language, 
\mr{consists in using} linguistic expressions that promote gender equality, inclusion of non-binary identities, and avoid reinforcing gender stereotypes \cite{rosola2023beyond}. 

In order to pursue the goals of \mr{fairness} and inclusiveness, measures that take into account the importance of the correlation between language and gender become central. 
Especially in heavily gender-marked languages such as Italian, the use and application of \mr{gender-fair} strategies is an urgent 
and yet difficult challenge. Indeed, in these languages, several are the elements one has to take into account to ensure a gender-fair use of language. However, adopting a gender-fair language is crucial given the negative effects of the masculine generics, documented in a range of empirical studies \cite{silveira1980generic,gygax2021challenge}; and \mr{recent years witnessed an increase in} 
awareness and effort to address these issues by promoting gender-fair language \cite{sulis-and-gheno-2022-the-debate}.

In Italian, the masculine is not only used to refer to and address men but also 
generic or unknown individuals; mixed-gender groups, regardless of the proportion of genders of its members; 
women, typically when occupying prestigious roles; and genderqueer people, given that there is no codified grammatical gender for referring to them \cite{gonzalez2024beyond}. This use, though, makes women and genderqueer people invisible, giving rise to a proper injustice \cite{rosola24inj, kapusta2016misgendering, dembroff2018he}. 
Extensive empirical literature also highlights how certain gendered expressions influence our cognition, with masculine terms evoking male images and reducing, e.g, the likelihood of women applying for or being considered suitable for a job position (for an overview see \cite{sczesny2016can, gygax2021cerveau}). 

Crucially, 
\ap{such unfair linguistic practices} \mr{are}
perpetuated in language technologies \cite{blodgett-etal-2020-language}. This becomes particularly evident in languages, like Italian, for which NLP tools often adopt masculine and stereotypical representations, making undue binary gender assumptions \cite{savoldi-etal-2024-prompt}.

We propose the \textbf{Gender-Fair Generation challenge} at CALAMITA 2024 \cite{calamita2024}, whose goal is to reduce the use of 
gender-unfair expressions in written Italian, focusing on both monolingual and cross-lingual scenarios (English-Italian). 
Our challenge is \textbf{structured into} \textbf{three tasks}---\textit{i)} gendered language detection, \textit{ii)} fair reformulation, and \textit{iii)} fair translation---\textbf{across three different datasets}. 
Namely, the newly created GFL-it corpus, composed of Italian texts extracted from 35 documents provided by the academic administration office of the University of Brescia and annotated 
following specific guidelines \cite{rosola2023beyond}; GeNTE, a bilingual test set for gender-neutral rewriting and translation built on a subset of the Europarl dataset \cite{piergentili-etal-2023-hi}; and Neo-GATE, a bilingual test set designed to evaluate the use of nonbinary neomorphemes in Italian
\cite{piergentili2024neomorphemes}.\footnote{\mr{In this report, we refer to innovative gender-fair strategies such as the schwa as ``neomorphemes". Although aware that this terminology is controversial, we adopted it for simplicity and do not intend our terminology to imply any substantive stance.}} 
We combine and repurpose these datasets across the three tasks envisioned in the Gender-Fair Generation challenge. 

This report is structured as follows: in Section \ref{sec:challenge-description}, we provide a description of our challenge; in Section \ref{sec:data_description}, we present the three datasets in detail; in Section \ref{sec:metrics}, we describe the metrics involved in our task; in Section \ref{sec:limitations}, we describe the limitations of our work, and finally, in Section \ref{sec:ethical-issues}, we 
\mr{discuss} the ethical issues.

\section{Challenge: Description}
\label{sec:challenge-description}

The Gender-Fair Generation challenge is organized into three tasks, which we present in detail below. \\

\noindent \textbf{1) Gendered language detection}: the first task tests the models’ ability to identify referentially gender-marked expressions within Italian sentences\mr{, namely those expressions whose (typically grammatical) gender is linked to their human referent}. Referentially gendered \mr{(henceforth 
simply \textit{gendered})} language includes:
    \begin{itemize}
    \item the \texttt{overextended} masculine or feminine, i.e., the use of a single \mr{gendered expression} 
    to refer to persons belonging to a mixed-gender group - e.g., \textit{i cittadini} (the.M citizens:M) used for a group of citizens of different genders;
    \item	the \texttt{generic} masculine or feminine, i.e., the use of a single \mr{gendered expression} 
    to refer to a generic or unknown person - e.g., \textit{il candidato deve avere tutti i requisiti} (the.M candidate:M has to possess all the requirements);
    \item	the \texttt{incongruous} gender, i.e., the use of a grammatical gender that does not match the referent's gender - e.g., \textit{il professore ordinario Maria Rossi} (the.M full.M professor:M Maria Rossi).
    \end{itemize}
    
\noindent \textbf{2) Fair reformulation}: the second task tests 
models’ ability to rewrite 
gendered expressions into alternative gender-fair expressions. To achieve this goal, various 
gender-fair language strategies can be employed.
In particular, we will employ \textit{obscuration} strategies:
    \begin{itemize}
    \item \textit{conservative obscuration}, i.e., the use of expressions and constructions that avoid providing information on the referent's gender  – e.g., \textit{il corpo docente} (the teaching body) or \textit{coloro che insegnano} (those who teach) instead of \textit{i professori} (the.M professors:M);
    \item \textit{innovative obscuration}, i.e., the use of novel, gender-neutral mark\mr{ers} 
    instead of the gendered ones – e.g., \textit{l{\textschwa} professor{\textschwa}} (the.INN professor:INN) instead of \textit{il professore} (the.M professor:M) or \textit{la professoressa} (the.F professor:F).\footnote{We indicate the innovative forms with ``INN" in the gloss\mr{es}.}
    \end{itemize}
    
As we further discuss in Section \ref{sec:data_description}, the
released version of 
GFL-it for this challenge and GeNTE 
\ap{include references and annotations designed for}
the former strategy, 
\ap{whereas Neo-GATE for the latter.}

\mr{
\apx{Note}
that} the chosen strategies do not exhaust the full range of possibilities\mr{:} we discarded, for the moment, \textit{visibility} strategies such as 
the repetition of an expression in the feminine and the masculine 
- e.g., \textit{i professori e le professoresse} (the.M professors:M and the.F professors:F) - 
and the repetition of \mr{in three gendered forms} 
(feminine, masculine \mr{and innovative)} 
– e.g., \textit{i professori, l{\textschwa} professor{\textschwa} e le professoresse} (the.M professors:M, the.INN professors:INN and the.F professors:F).


\bigskip
\noindent \textbf{3) Fair translation}: like the second task, the third one is designed to test the models' ability to generate gender-fair language texts, but in the \textit{cross-lingual} context of \textit{automatic translation} from English into Italian. 
%
%
For example, consider applying the two gender-fair language strategies described above to the translation of the 
sentence ``I am glad to know such knowledgeable doctors'':

\begin{itemize}
    \item conservative obscuration: \textit{Sono felice di conoscere un \ul{personale medico} così preparato.} \ap{\textsubscript{[medical staff]}}
    \item innovative obscuration: \textit{Sono content\underline{{\textschwa}} di conoscere medic\underline{{\textschwa}} così preparat\underline{{\textschwa}}.}
\end{itemize}


\section{Data description}
\label{sec:data_description}

\setlength\tabcolsep{2.4pt}
\begin{table}
    \centering
    \begin{tabular}{lcccc}
    \toprule
        \textbf{Task} & \textbf{GFL-it} &\textbf{GeNTE} & \textbf{Neo-GATE} & \textbf{Task total}\\ \midrule
        \textbf{Detection} & 2,170 & - & 841 & 3,011 \\
         \textbf{Reformulation} & 1,215 & 750 & 841 & 2,806\\
         \textbf{Translation} & - & 1,500 & 841 & 2,341\\
         \bottomrule
    \end{tabular}
    \caption{Number of dataset entries used for each task.}
    \label{tab:stats}
\end{table}

For our challenge\mr{,} we propose three benchmarks dedicated to the evaluation of gender-fair language generation\mr{, (}
GFL-it\footnote{\url{https://github.com/simonasnow/GFL-it-Dataset}}, GeNTE \cite{piergentili-etal-2023-hi},\footnote{\url{https://huggingface.co/datasets/FBK-MT/GeNTE}} and Neo-GATE \cite{piergentili2024neomorphemes}\mr{)},\footnote{\url{https://huggingface.co/datasets/FBK-MT/Neo-GATE}} and a total of 7 prompts to be used across the tasks and 
datasets.
We describe 
the datasets
in \mr{sub}sections \ref{sec:gfl-it}, \ref{sec:gente}, and \ref{sec:neo-gate} respectively, and the prompts in \mr{sub}section \ref{sec:prompts}. 

Statistics about the benchmarks and their use within this challenge proposal are available in Table \ref{tab:stats}.
GFL-it contains 
\mr{a} total \mr{of} 2,170 texts, 
\mr{among} which 5 \mr{expert} annotators identified 
an average of 
\apx{3.54}
unfair spans (in total 4,311) in 1,215 texts. 
The annotators proposed on average 1.10 conservative obscuration alternatives.

%
For more detailed statistics about GeNTE and Neo-GATE we refer to the respective papers.




\subsection{GFL-it}
\label{sec:gfl-it}

GFL-it was built on documents and texts from University website pages provided by the University of Brescia. It constitutes an expansion of the corpus presented in 
\ap{\citet{rosola2023beyond}.}
The corpus comprises a total of 35 documents in Italian, split into 2,187 texts. Each text was annotated by 5 \last{paid} expert annotators following the \ap{original} annotation scheme 
\cite{rosola2023beyond}. First, 
\ap{the annotators}
identified all the spans that contained any gender-unfairness, distinguishing among: \textsc{overextended} (3,772), \textsc{generic} (597)\footnote{Some spans (e.g., `alcuni docenti') have been identified by annotators as overextended and generic type of gendered language.} and \textsc{incongruous gender} (31) (see \ref{sec:challenge-description}). 
Then, they provided at least one alternative per span. The alternatives could belong to any \ap{of the} gender-fair strategies:
conservative or innovative obscuration, conservative or innovative visibility,
or hybrid alternatives (i.e., any combination of these types). 

\mr{Given that} GFL-it is 
annotated for spans, 
each text contains a list of different spans and their reformulations in different forms of \mr{gender-fair} language\footnote{For the purpose of the task 2, only the conservatively obscured reformulations have been released in this version of the dataset.}. 
More specifically, each entry is described by the following attributes:
\begin{itemize}
    \item \textit{id\_text}: The unique ID for each text.
    \item \textit{text}: The entire text of the entry.
    \item \textit{label}: This label is used for task 2: if the text contains gendered expressions and conservative obscuration alternatives have been proposed by annotators, the label is \textsc{0}; if the text contains gendered expressions but no conservative obscuration alternatives have been proposed, the label is \textsc{1}; otherwise if no gendered expressions have been detected in the text, the label field is empty.
    \item \textit{list\_spans}: The list containing all spans found in the text.
    
\end{itemize}

Each span in \textit{list\_spans} follows the structure:
\begin{itemize}
    \item \textit{span}: The textual representation of the span.
    \item \textit{start}: The starting index of the span in the text.
    \item \textit{end}: The ending index of the span in the text.
    \item \textit{types\_of\_gendered\_language}: A list of the 
    types of 
    gendered \mr{language used in} 
    the selected spans
    \mr{;} 
    \apx{possible values are \textsc{overextended}, \textsc{generic} and \textsc{incongruous gender}.}
    \item \textit{key\_span}: The concatenation of \textit{span}, \textit{start} and \textit{end} \mr{attributes;} it can be used as an ID for each span contained inside a text.
    \item \textit{AOC}: The list of conservative obscuration alternatives proposed by annotators.
\end{itemize}

\mr{We propose to use the} GFL-it corpus 
\mr{for} tasks 1 and 2,\footnote{\mr{For task 2, we used a classifier that distinguishes between gendered and gender-neutral texts (see Section \ref{sec:metrics}). Hence,} 
we \ap{only} used
the \mr{GFL-it} texts \mr{where the annotators identified} 
gendered expressions 
and 
provided at least one conservatively obscured reformulation
(= \mr{gender}-neutral class) for a total amount of 1,120 texts.} namely, those regarding \textbf{gendered language detection} and \textbf{fair reformulation}.

\setlength\tabcolsep{5pt}
\begin{table*}[htp!]
\centering
    \begin{tabularx}{\textwidth}{lX}
    
    \toprule
    \textbf{Text} & Per \textbf{gli iscritti }agli anni successivi al primo tali valutazioni scendono rispettivamente a  NUM ,  NUM  (sotto la soglia critica) e  NUM  (vicino alla soglia critica). \\
    \textbf{Span} & gli iscritti \\
    \textbf{\mr{Reformulated} Text} & Per \textbf{le persone iscritte} agli anni successivi al primo tali valutazioni scendono rispettivamente a  NUM ,  NUM  (sotto la soglia critica) e  NUM  (vicino alla soglia critica).\\
    & \apx{\textsubscript{[For 
\mr{those} enrolled in years after the first, these ratings drop to NUM, NUM (below the critical threshold) and NUM (close to the critical}} \apx{\textsubscript{threshold), respectively.]}} \\
    \bottomrule
    \end{tabularx}
\caption{Example from the GFL-it dataset. 
\mr{Words in bold correspond to the identified unfair spans in the text, and the reformulated expressions in the reformulated text.}
\apx{A translation of the text is provided in square brackets.}
}
\label{tab:gfl-it_examples}
\end{table*}

\setlength\tabcolsep{5pt}
\begin{table*}[htp!]
\centering
    \begin{tabularx}{\textwidth}{clX}
    \toprule
    \multirow{4}{*}{\textbf{Set-G}} & \textbf{SRC} & When you assumed office, \ul{Mr} Schreyer, you assured us that you would strive to achieve this.\\
     & \textbf{REF-G} & Al momento della sua nomina, \textbf{signor} \textsubscript{[Mr]} Schreyer, ci aveva promesso che si sarebbe \textbf{adoperato} \textsubscript{[(would have) strived]} in tal senso. \\
      & \textbf{REF-N} & Al momento della sua nomina, Schreyer, ci aveva promesso \textit{un impegno} \textsubscript{[a commitment]} in tal senso. \\\midrule
    \multirow{4}{*}{\textbf{Set-N}} & \textbf{SRC} & To some extent, those of us who are politicians find ourselves in the middle. \\
     & \textbf{REF-G} & In certa misura \textbf{quelli} \textsubscript{[those (of us)]} di noi che sono \textbf{politici}\textsubscript{[politicians]} si trovano in una posizione intermedia.\\
      & \textbf{REF-N} & In certa misura \textit{chi} di noi \textsubscript{[who, among us,]} \textit{svolge attività politica} \textsubscript{[carries out political activities]} si trova in una posizione intermedia. \\\midrule
    \end{tabularx}
\caption{Examples of Set-G and Set-N entries in GeNTE. \ul{Underlined} words are linguistic cues informing about human referents' gender; words in \textbf{bold} are gendered mentions of human referents; words in \textit{italic} are the \mr{gender-}neutral reformulations of the gendered 
\mr{mentions}. Glosses of relevant expressions are 
\mr{provided} in 
square brackets.}
\label{tab:gente_examples}
\end{table*}

\setlength\tabcolsep{5pt}
\begin{table*}[htp!]
\centering
    \begin{tabularx}{\textwidth}{lX}
    \toprule
    \textbf{SOURCE} & After the accident, they took me to the hospital and I stayed there for a whole month. \\
    \textbf{REF-M} & Dopo l'incidente, mi hanno portato all'ospedale e sono rimasto lì per un mese intero. \\
    \textbf{REF-F} & Dopo l'incidente, mi hanno portata all'ospedale e sono rimasta lì per un mese intero. \\
    \textbf{REF-TAGGED} & Dopo l'incidente, mi hanno \ul{portat{\textschwa}} all'ospedale e sono \ul{rimast{\textschwa}} lì per un mese intero. \\
    \textbf{ANNOTATION} & portato portata \ul{portat{\textschwa}}; rimasto rimasta \ul{rimast{\textschwa}}; \\
    \bottomrule
    \end{tabularx}
\caption{Example of a Neo-GATE entry, already adapted to the \texttt{schwa-simple} neomorpheme paradigm. \ul{Underlined} words include the neomorpheme schwa (\textschwa).}
\label{tab:neogate_examples}
\end{table*}



\subsection{GeNTE}
\label{sec:gente}
GeNTE is a parallel English\mr{~→~}Italian test set \citep{piergentili-etal-2023-gender}. 
Originally designed to evaluate MT models' ability to perform gender-neutral translations, 
GeNTE was built upon a subset of the Europarl corpus \citep{koehn2005europarl}, which  
is representative of natural, formal communicative situations from the institutional domain, the context where gender-neutral language is most accepted and encouraged \cite{piergentili-etal-2023-gender, piergentili-etal-2023-hi}. 
Overall, it consists of 1,500  <\textit{English source, gendered Italian reference, \mr{gender-}neutral Italian reference}> triplets aligned at the sentence level, which always contain at least one mention 
of
human referents.
The gendered Italian reference (REF-G) comes from the original Europarl corpus, whereas the \mr{gender-}neutral reference (REF-N) was produced by professional translators 
\mr{who} edited gendered forms into \mr{gender-}neutral alternatives.

As shown in Table \ref{tab:gente_examples}, GeNTE  represents two types of phenomena, which are equally represented within the corpus. Namely, \textit{i)} \textsc{Set-N}, featuring 750 
gender-ambiguous source 
sentences that require to be rendered \mr{gender-}neutrally; and \textit{ii)}
\textsc{Set-G}
featuring gender-unambiguous source sentences, to
be properly rendered with gendered (masculine or feminine) forms. 
Crucially, these two sets are a key feature of GeNTE, as they allow benchmarking whether systems are able to perform gender-neutral translations, but only when desirable. As a matter of fact,  when referents' gender is unknown or irrelevant, undue gender inferences should not
be made and \mr{gender-}neutral language (i.e., conservative obscuration strategy) should be used. However, \mr{gender-}neutralization should not be always enforced, and when a referent's gender is known or relevant, models should not over-generalize to \mr{gender-}neutral generations.

Each entry in GeNTE is organized into the following fields:

\begin{itemize}
    \item \textit{ID}: The unique GeNTE ID.
    \item  \textit{Europarl\_ID}: The original sentence ID from Europarl's common-test-set 2.
    \item \textit{SET}: Indicates whether the entry belongs to the \textsc{Set-G} or the \textsc{Set-N} subportion of the corpus.
 \item \textit{SRC}: The English source sentence.
 \item \textit{REF-G}: The gendered Italian reference translation.
 \item \textit{REF-N}: The gender-neutral Italian reference, produced by a professional translator. 
 \item \textit{GENDER}: For entries belonging to the Set-G, it indicates if the entry is Feminine or Masculine
 .

\end{itemize}

We propose the use of the whole GeNTE 
for the \textbf{translation }\textbf{task 3}, testing models' ability to produce \mr{gender-}neutral translations only when appropriate. For the \textbf{fair reformulation} \textbf{task 2}, we only repurpose \apx{part of} the Italian portion of the corpus, 
\apx{i.e., REF-G references from}
\textsc{Set-N}
.




\subsection{Neo-GATE}
\label{sec:neo-gate}
Similarly to GeNTE, Neo-GATE is a parallel corpus designed for gender-\mr{fair} English~→~Italian MT evaluation. Here, however, the focus is on the use of gender-\mr{fair} neomorphemes (i.e., innovative obscuration strategy) rather than conservative gender-neutral language.
Neo-GATE was built on GATE \cite{rarrick2023gate}, a test set \mr{manually} 
created \mr{specifically} to evaluate gender reformulation and gender bias in MT. In GATE, the gender of human entities is unknown, i.e., there are no linguistic elements providing gender information about human referents in the (English) source sentences.

Neo-GATE includes an annotation that defines the words upon which the evaluation is based. It includes the three forms required for the evaluation, i.e., the masculine and feminine forms, and forms featuring placeholders in place of Italian overt gender markers. 
Before the evaluation, the placeholders must be replaced with the correct forms in the desired neomorpheme paradigm. For this task, Neo-GATE was adapted to a version of the `schwa' paradigm \cite{thornton_genere_2020,baiocco_italian_2023}, to which we refer as \texttt{schwa-simple} here, i.e., the placeholders were replaced with the forms described in Appendix \ref{appendix:tagset}. 

Like GeNTE, Neo-GATE includes Italian references that differ exclusively in gender expression. Besides the English source sentence, all entries in Neo-GATE have three Italian references: REF-M, where the gender of words referring to human beings is masculine, REF-F, where human beings are referred to as feminine, and REF-TAGGED, where placeholders replace overt markers of gender 
-- here adapted to the \texttt{schwa-simple} paradigm. However, differently from GeNTE, the English sentences in Neo-GATE never include gender cues. An example of a Neo-GATE entry is available in Table \ref{tab:neogate_examples}.

Each entry in Neo-GATE includes the following fields:

\begin{itemize}
    \item \textit{\#}: The entry identifier within Neo-GATE.
    \item \textit{GATE-ID}: A unique identifier of the original GATE entry, composed of a prefix indicating the subset of origin followed by a serial number.
    \item \textit{SOURCE}: The English source sentence.
    \item \textit{REF-M}: The Italian reference where all gender-marked terms are masculine.
    \item \textit{REF-F}: The Italian reference where all gender-marked terms are feminine.
    \item \textit{REF-TAGGED}: The Italian reference where all gender-marked terms are tagged with Neo-GATE's annotation.
    \item \textit{ANNOTATION}: The word level annotation.

\end{itemize}

We propose to use all Neo-GATE entries for \textbf{all three tasks} of our challenge. While for tasks 1 (\textbf{gendered language detection}) and 2 (\textbf{fair reformulation}) we only use Italian references -- namely \ap{
both REF-M and REF-F for task 1, 
and REF-M only 
for task 2} 
-- as input for the models, for task 3 (\textbf{fair translation}) we use the English SOURCE sentences.









\setlength\tabcolsep{3.5pt}
\begin{table*}[ht]
\centering
    \begin{tabularx}{\textwidth}{llX}
    \toprule
    \multicolumn{3}{c}{\textbf{Task 1: Gendered language detection}} \\
    \midrule
    \multirow{1}{*}{\textbf{A}} & 
    \textbf{Instruction} & Identifica le espressioni che contengono dei marcatori di genere femminile o maschile. Se più di un'espressione è identificata, separale con \mr{`` ;}". Se nessuna espressione è identificata, restituisci 0.
    \\
    & {\textbf{Exemplar format}} & [Genere marcato]: A livello di Ateneo ha partecipato il 76\% degli  iscritti del 1° anno \mr{del} CdL e l’81\% di tutti gli altri studenti.\\
    & & [Espressione]: degli iscritti ; 
    tutti gli altri studenti 
    \\
    \midrule
    
    \multirow{1}{*}{\textbf{B}} & 
    \textbf{Instruction} & Identifica le espressioni che contengono dei marcatori di genere femminile o maschile. Se più di un'espressione è identificata, separale con \mr{`` };". Se nessuna espressione è identificata, restituisci 0.\\
    & {\textbf{Exemplar format}} & [Genere marcato]: Quest’anno mi sono ammalata già due volte.\\
    &&[Espressione]: ammalata
    \\
    \midrule
    
    \multicolumn{3}{c}{\textbf{Task 2: Fair reformulation}} \\
    \midrule
    \multirow{1}{*}{\textbf{C}} &  
    \textbf{Instruction} & Riformula la seguente frase utilizzando un linguaggio neutro rispetto al genere dei referenti umani, evitando l’uso di forme maschil\mr{i} e femminili.\\
    & {\textbf{Exemplar format}} & [Genere marcato]: - Il 79\% dei laureati lavora (a un anno dalla laurea).\\
    &&[Neutro]: - Il 79\% delle persone laureate lavora (a un anno dalla laurea). \\
    \midrule
    
    \multirow{6}{*}{\textbf{D}} & \textbf{Instruction} & Riformula la seguente frase utilizzando un linguaggio neutro rispetto al genere dei referenti umani, evitando l’uso di forme maschil\mr{i} e femminili. \\
    & {\textbf{Exemplar format}} & [Genere marcato]: Secondariamente, fino a che punto aumenta la trasparenza e la responsabilità dei parlamentari europei? \\
    & & [Neutro]: Secondariamente, fino a che punto aumenta la trasparenza e la responsabilità dei membri del Parlamento Europeo?\\
    \midrule
    \multirow{4}{*}{\textbf{E}} & \textbf{Instruction} & Riformula la seguente frase usando il neomorfema ``\textschwa". Il neomorfema ``\textschwa" deve essere usato come sostituto dei morfemi maschili e femminili nelle parole che si riferiscono agli esseri umani. \\
    & {\textbf{Exemplar format}} & [Genere marcato]: Il partner di un mio amico ci ha invitati a cena. \\
    & & [Neomorfema]: L{\textschwa} partner di un{\textschwa} mi{\textschwa} amic{\textschwa} ci ha invitat{\textschwa} a cena.\\
    \midrule
    \multicolumn{3}{c}{\textbf{Task 3: Fair translation}} \\
    \midrule
    \multirow{12}{*}{\textbf{F}} & \textbf{Instruction} & Traduci la seguente frase inglese in italiano seguendo queste regole:\\
    & & 1. Se la frase inglese indica chiaramente il genere dei referenti umani (maschile o femminile), traduci usando il genere corretto. \\
    & & 2. Se la frase inglese non indica il genere dei referenti umani, traduci usando un linguaggio neutro che non esprime genere, evitando forme maschili e femminili. \\
    & {\textbf{Exemplar format 1}} & [Inglese]: However, it is important that the Commissioner has declared his loyalty to the President himself. \\
    & & [Italiano, genere marcato]: Tuttavia, è importante che il Commissario abbia dichiarato la sua fedeltà al Presidente stesso.\\
    & {\textbf{Exemplar format 2}} & [Inglese]: Secondly, how far does it increase transparency and accountability of the MEPs? \\
    & & [Italiano, neutro]: Secondariamente, fino a che punto aumenta la trasparenza e la responsabilità dei membri del Parlamento Europeo?\\
    \midrule
    \multirow{4}{*}{\textbf{G}} & \textbf{Instruction} & Traduci la seguente frase inglese in italiano usando il neomorfema ``{\textschwa}". Il neomorfema ``{\textschwa}" deve essere usato come sostituto dei morfemi maschili e femminili nelle parole che si riferiscono agli esseri umani. \\
    & {\textbf{Exemplar format}} & [Inglese]: The partner of a friend of mine invited us to dinner. \\
    & & [Italiano, genere marcato]: Il partner di un mio amico ci ha invitati a cena.\\
    & & [Italiano, neomorfema]: L{\textschwa} partner di un{\textschwa} mi{\textschwa} amic{\textschwa} ci ha invitat{\textschwa} a cena. \\
    \bottomrule
    \end{tabularx}
    \caption{Examples of the format of all prompts we propose for our challenge. 
    Dataset-wise, prompts A and C are designed to be used with GFL-it data, prompts B, E, and G are designed for Neo-GATE, and prompts D and F are designed for GeNTE.}
    \label{tab:prompts}
\end{table*}

\subsection{Example of used prompts}
\label{sec:prompts}

This section describes 
\ap{the}
prompts we propose for our challenge
\apx{, with examples available in}
Table \ref{tab:prompts}.

In prompts \textbf{A} 
and \textbf{B}
, we ask the model to identify the 
gendered expressions (introduced by the \ap{tag} \textit{[Espressione]:})
 in the text given as input; if no 
 gendered expression is detected in the text (initialized with the tag \textit{[Genere marcato]:}) the model should output 0. The model can recognize more than one 
 gendered expression. 

In prompts \textbf{C}
, \textbf{D}
, and \textbf{E}, the shots include one line starting with the tag \textit{[Genere marcato]:}, indicating that the following sentence is 
gendered. Then, in prompts \textbf{C} and \textbf{D} the following line starts with \textit{[Neutro]:} followed by a gender-neutral reformulation, whereas in \textbf{E} it starts with \textit{[Neomorfema]:} and includes the innovative obscuration alternative 
of the first sentence, with neomorphemes in place of the masculine forms.\footnote{\mr{We here used \textit{neutro} (neutral/neuter), despite being aware of its ambiguity with \textit{neuter}, a grammatical gender not present in the Italian linguistic system. 
However, nothing substantive hinges on this terminological choice.
}}

Prompts \textbf{F} and \textbf{G} start with the tag \textit{[Inglese]:} followed by the English source sentence to be translated. In prompt \textbf{F}, the second line either starts with the tag \textit{[Italiano, genere marcato]:} (see F - Exemplar format 1 in Table \ref{tab:prompts}) if it is followed by a 
gendered translation or with the tag \textit{[Italiano, neutro]:} if the subsequent translation is gender-neutral (see F - Exemplar format 2). Models are required to produce the correct tag and translation depending on the presence or absence of gender cues in the source. Finally, prompt \textbf{G} includes two different translations after the source sentence: the first, preceded by the tag \textit{[Italiano, genere marcato]:}, includes a translation featuring masculine forms in reference to human beings, whereas the second translation starts with the tag \textit{[Italiano, neomorfema]:} and uses neomorphemes in reference to human beings. Models are required to produce both translations, though only the second will be extracted in post-processing and used for the evaluation.

In particular, prompts D, E, F, and G are based on the ones used in previous experiments on the same datasets \cite{savoldi-etal-2024-prompt,piergentili2024neomorphemes}, and were in turn inspired by the format proposed by \citet{sanchez2024genderspecific}.



\section{Metrics}
\label{sec:metrics}




For the evaluation of gendered language detection 
\mr{(}i.e., with GFL-it and Neo-GATE in task 1\mr{)} 
we used the F1-score obtained using BERTScore\footnote{\url{https://huggingface.co/spaces/evaluate-metric/bertscore}} \cite{bert-score} for each entry in the datasets. In particular, for each entry, we extract the most relevant correspondence between the 
gendered expressions identified by the annotators \mr{and} 
the ones produced by the generative model, computing the maximum F1-score. 
Once the correspondences are set for each entry, we average the scores.

For the evaluation of gender-neutral reformulation---i.e., with GFL-it and GeNTE in task 2---and translation---i.e., with GeNTE \mr{and Neo-GATE} in task 3---we propose an accuracy score based on the labels produced by \mr{the} classifier introduced in \citet{piergentili-etal-2023-hi}. More specifically, we use version 2 of the classifier, introduced in \citet{savoldi-etal-2024-prompt}. Th\mr{is} classifier assigns a label to each model output, either \mr{gender-}neutral or 
gendered. We then compare those labels against the true labels, i.e., always \mr{gender-}neutral in the reformulation task and either gendered or \mr{gender-}neutral for the translation task, depending on whether the entry belongs to \textsc{Set-G} or \textsc{Set-N} respectively. The final score is computed as the corpus-level percentage of correct labels. 

For neomorpheme-based gender-\mr{fair} reformulation (task 2) and translation (task 3) based on Neo-GATE, we propose the coverage-weighted accuracy described in \citet{piergentili2024neomorphemes} as the main metric. This metric takes into account both how accurately a model generates neomorphemes and the proportion of annotations (i.e., either of the masculine, feminine, or innovative forms) found during the evaluation, thus allowing for fair system comparisons and rankings.
As complementary metric to assess models' ability to correctly generate neomorphemes, we propose reporting the mis-generation score \cite{piergentili2024neomorphemes} as well. This metric can 
\mr{flag} undesired behaviors even despite good accuracy, as it counts cases where models generate neomorphemes inappropriately, for instance by applying the use of neomorphemes to words that do not refer to human entities (e.g., by generating \mr{‘}tavol{\textschwa}\mr{'} instead of ‘tavolo’, en: table).

\section{Limitations}
\label{sec:limitations}

Our work presents some limitations. Firstly, the datasets employed only derive from specific domains: GFL-it exclusively contains data from administrative documents \mr{and official web pages of the University}, GeNTE from documents of the European Parliament, and Neo-GATE data manually created by experts. The corpora could be expanded \mr{to other domains} and annotated by more annotators in future research. Secondly, our metrics are only a first attempt and others should be explored in the future. Moreover, we only tested one paradigm of neomorphemes, namely the \texttt{schwa-simple}, while many others exist (e.g., the asterisk, the ‘-u', the ‘$@$' - see \cite{ghenoFalla} for a complete list), and even more could be proposed. Furthermore, GeNTE and Neo-GATE do not contain mixed texts where rewriting is needed with respect to one entity but not others.


\section{Ethical issues}
\label{sec:ethical-issues}


The proposed tasks in this challenge have the purpose of reducing the use of gender\mr{-unfair} expressions in \mr{heavily} gender-marked languages (i.e., Italian) that affect the visibility of other genders (in particular, feminine and non-binary). Although the datasets have been built by experts of \mr{gender-fair} language, the 
group of annotators \mr{of GFL-it} was not \mr{gender-}balanced 
as \mr{only 2} out of \mr{5} annotators were \mr{men}. 

Moreover, we are aware of the fact that the use of neomorphemes like the schwa {\textschwa} makes reading harder for people with dyslexia or visual impairments \cite{sulis-and-gheno-2022-the-debate,Iacopini-2021-schwa-inaccessibile,DeSantis-2022-emancipazione-rovesciata}. This issue, however, is mitigated thanks to the possibility of selecting the most suitable neomorpheme according to each user's needs. \mr{In particular, both people with dyslexia or visual impairments can rely on screen readers, which differ in their ability to correctly interpret specific neomorphemes: the possibility to select different \apx{neo}morphemes allows each user to select the one(s) their screenreader interpret best.}

\section{Data license and copyright issues}

Creative Commons Attribution 4.0 International license (CC BY 4.0).
\url{https://creativecommons.org/licenses/by-sa/4.0/deed.it}

\section{Acknowledgments}

Beatrice Savoldi 
is supported by the PNRR project FAIR -  Future AI Research (PE00000013),  under the NRRP MUR program funded by the NextGenerationEU. Luisa Bentivogli is funded by the Horizon Europe research and innovation programme,  under grant agreement No 101135798, project Meetween (My Personal AI Mediator for Virtual MEETtings BetWEEN People). The work of Viviana Patti and Marco Madeddu is supported by “HARMONIA” project - M4-C2, I1.3 Partenariati Estesi - Cascade Call - FAIR - CUP C63C22000770006 - PE PE0000013 under the NextGenerationEU programme.

The annotation of GFL-it has been partially funded by Università degli Studi di Brescia as part of the actions provided for by the Gender Equality Plan.








\bibliography{sample-ceur,custom}

\appendix

\section{The \texttt{schwa-simple} paradigm}
\label{appendix:tagset}

Table \ref{tab:tagset} reports the forms used in the \texttt{schwa-simple} paradigm, along with the corresponding tags in Neo-GATE and masculine and feminine equivalents.

\setlength{\tabcolsep}{2pt}
\begin{table*}[t]
\small
\centering
\begin{tabular}{lllll}
\toprule
\textbf{TAG}& \textbf{Description} & \textbf{Masculine}&  \textbf{Feminine} & \textbf{Schwa} \\
\midrule
\textless{}ENDS\textgreater{}& \begin{tabular}[c]{@{}p{7.2cm}@{}} \mr{portion of the word differentiating 
gendered forms}
, singular \end{tabular}  & o, e, tore& a, essa, trice  & \textschwa, tor\textschwa~ \\
\textless{}ENDP\textgreater{}& \begin{tabular}[c]{@{}p{7.2cm}@{}} \mr{portion of the word differentiating 
gendered forms}
, plural \end{tabular} & i, tori& e, esse, trici & \textschwa, tor\textschwa~   \\
\textless{}DARTS\textgreater{}& \begin{tabular}[c]{@{}p{6.5cm}@{}} definite article, singular \end{tabular}  & il, lo, l'& la, l' & l\textschwa~ \\
\textless{}DARTP\textgreater{}&  \begin{tabular}[c]{@{}p{6.5cm}@{}} definite article, plural \end{tabular} &i, gli& le & l\textschwa~\\
\textless{}IART\textgreater{}& \begin{tabular}[c]{@{}p{6.5cm}@{}} indefinite article \end{tabular} & uno, un& una, un' & un\textschwa~\\
\textless{}PARTP\textgreater{}& \begin{tabular}[c]{@{}p{6.5cm}@{}} partitive article, plural \end{tabular} & dei, degli& delle& de\textschwa~ \\
 
\textless{}PREPdiS\textgreater{}& \begin{tabular}[c]{@{}p{6.5cm}@{}} articulated preposition with root `di', singular\end{tabular}  & del, dello, dell'& della, dell' &  dell\textschwa~     \\
\textless{}PREPdiP\textgreater{}& \begin{tabular}[c]{@{}p{6.5cm}@{}} articulated preposition with root `di', plural\end{tabular} & dei, degli& delle& dell\textschwa~    \\
\textless{}PREPaS\textgreater{}& \begin{tabular}[c]{@{}p{6.5cm}@{}} articulated preposition with root `a', singular\end{tabular} & al, allo, all'& alla, all'&  all\textschwa~\\
\textless{}PREPaP\textgreater{}& \begin{tabular}[c]{@{}p{6.5cm}@{}} articulated preposition with root `a', plural\end{tabular} & agli, ai& alle& all\textschwa~\\
\textless{}PREPdaS\textgreater{}& \begin{tabular}[c]{@{}p{6.5cm}@{}} articulated preposition with root `da', singular\end{tabular}  & dal, dallo, dall'& dalla, dall'& dall\textschwa~     \\
\textless{}PREPdaP\textgreater{}& \begin{tabular}[c]{@{}p{6.5cm}@{}} articulated preposition with root `da', plural\end{tabular}  & dagli& dalle& dall\textschwa~     \\
\textless{}PREPinP\textgreater{}& \begin{tabular}[c]{@{}p{6.5cm}@{}} articulated preposition with root `in', plural\end{tabular}  & negli& nelle&  nell\textschwa~    \\
\textless{}PREPsuS\textgreater{}& \begin{tabular}[c]{@{}p{6.5cm}@{}} articulated preposition with root `su', singular\end{tabular}  & sul, sullo, sull'& sulla, sull'&  sull\textschwa~     \\
\textless{}PREPsuP\textgreater{}& \begin{tabular}[c]{@{}p{6.5cm}@{}} articulated preposition with root `su', plural\end{tabular} & sugli& sulle&  sull\textschwa~     \\
\textless{}DADJquelS\textgreater{}& \begin{tabular}[c]{@{}p{6.5cm}@{}} demonstrative adjective (far), singular\end{tabular} & quel, quello, quell'& quella, quell' & quell\textschwa~    \\
\textless{}DADJquelP\textgreater{}& \begin{tabular}[c]{@{}p{6.5cm}@{}} demonstrative adjective (far), plural\end{tabular}  & quegli& quelle & quell\textschwa~    \\
\textless{}DADJquestS\textgreater{}&  \begin{tabular}[c]{@{}p{6.5cm}@{}} demonstrative adjective (near), singular\end{tabular} & questo, quest'& questa, quest'& quest\textschwa~     \\
\textless{}DADJquestP\textgreater{}& \begin{tabular}[c]{@{}p{6.5cm}@{}} demonstrative adjective (near), plural\end{tabular}  & questi& queste& quest\textschwa~    \\
\textless{}POSS1S\textgreater{}& \begin{tabular}[c]{@{}p{6.5cm}@{}} possessive adjective, 1st person singular, singular\end{tabular}  & mio& mia & mi\textschwa~ \\
\textless{}POSS1P\textgreater{}& \begin{tabular}[c]{@{}p{6.5cm}@{}} possessive adjective, 1st person singular, plural\end{tabular}  & miei& mie&  mi\textschwa~ \\
\textless{}POSS2S\textgreater{}& \begin{tabular}[c]{@{}p{6.5cm}@{}} possessive adjective, 2nd person singular, singular\end{tabular} & tuo& tua & tu\textschwa~ \\
\textless{}POSS2P\textgreater{}& \begin{tabular}[c]{@{}p{6.5cm}@{}} possessive adjective, 2nd person singular, plural\end{tabular} & tuoi& tue& tu\textschwa~ \\
\textless{}POSS3S\textgreater{}&  \begin{tabular}[c]{@{}p{6.5cm}@{}} possessive adjective, 3rd person singular, singular\end{tabular} & suo& sua& su\textschwa~ \\
\textless{}POSS3P\textgreater{}& \begin{tabular}[c]{@{}p{6.5cm}@{}} possessive adjective, 3rd person singular, plural\end{tabular} & suoi& sue& su\textschwa~ \\
\textless{}POSS4S\textgreater{}& \begin{tabular}[c]{@{}p{6.5cm}@{}} possessive adjective, 1st person plural, singular\end{tabular}  & nostro& nostra & nostr\textschwa~     \\
\textless{}POSS4P\textgreater{}& \begin{tabular}[c]{@{}p{6.5cm}@{}} possessive adjective, 1st person plural, plural\end{tabular} & nostri& nostre& nostr\textschwa~     \\
\textless{}PRONDOBJS\textgreater{}& \begin{tabular}[c]{@{}p{6.5cm}@{}} direct object pronoun, singular \end{tabular} & lo& la & l\textschwa~  \\
\textless{}PRONDOBJP\textgreater{}& \begin{tabular}[c]{@{}p{6.5cm}@{}} direct object pronoun, plural \end{tabular} & li& le& l\textschwa~  \\
\bottomrule
\end{tabular}
\caption{The full tagset used in \textsc{Neo-GATE}, mapped to the Italian 
gendered forms and the \texttt{schwa-simple} nomorpheme paradigm.}
\label{tab:tagset}
\end{table*}



\end{document}